\documentclass{article}
\usepackage{iclr2026_conference,times}

\usepackage{amsmath,amsfonts,bm}

\def\eqref#1{equation~\ref{#1}}

\def\1{\bm{1}}

\DeclareMathAlphabet{\mathsfit}{\encodingdefault}{\sfdefault}{m}{sl}
\SetMathAlphabet{\mathsfit}{bold}{\encodingdefault}{\sfdefault}{bx}{n}

\usepackage{hyperref}
\usepackage{url}
\usepackage{enumitem}
\usepackage{algorithm}
\usepackage{algorithmic}
\usepackage{tikz}
\usepackage{graphicx}

\title{The Last Harness You'll Ever Build}

\author{Haebin Seong \\
Sylph.AI \\
\And
Li Yin \\
Sylph.AI \\
\And
Haoran Zhang \\
Sylph.AI \\
\And
Zhan Shi \\
Sylph.AI \\
}

\iclrfinalcopy 
\begin{document}

\maketitle

\begin{abstract}
AI agents are increasingly deployed on complex, domain-specific workflows---navigating enterprise web applications that require dozens of clicks and form fills, orchestrating multi-step research pipelines that span search, extraction, and synthesis, automating code review across unfamiliar repositories, and handling customer escalations that demand nuanced domain knowledge. \textbf{Each new task domain requires painstaking, expert-driven harness engineering}: designing the prompts, tools, orchestration logic, and evaluation criteria that make a foundation model effective. We present a two-level framework that automates this process. At the first level, the \textbf{Harness Evolution Loop} optimizes a worker agent's harness $\mathcal{H}$ for a single task: a Worker Agent $W_{\mathcal{H}}$ executes the task, an Evaluator Agent $V$ adversarially diagnoses failures and scores performance, and an Evolution Agent $E$ modifies the harness based on the full history of prior attempts. At the second level, the \textbf{Meta-Evolution Loop} optimizes the evolution blueprint $\Lambda = (W_{\mathcal{H}}, \mathcal{H}^{(0)}, V, E)$ itself across diverse tasks, \textbf{learning a blueprint $\Lambda^{(\text{best})}$ that enables rapid harness convergence on any new task---so that adapting an agent to a novel domain requires no human harness engineering at all.} We formalize the correspondence to meta-learning and present both algorithms. The framework \textbf{shifts manual harness engineering into automated harness engineering}, and takes one step further---\textbf{automating the design of the automation itself}.
\end{abstract}

\section{Introduction}
\label{sec:intro}

Recent work on \emph{harness engineering} has demonstrated that carefully designed scaffolding---execution environments, feedback loops, evaluation criteria, and context management---can \textbf{dramatically amplify} what agents achieve \citep{lopopolo2026harness, rajasekaran2026harness}. However, these harnesses are themselves products of \textbf{highly intensive, specialized human engineering}. \citet{lopopolo2026harness} describe building custom linters, repository-local observability stacks (logs, metrics, traces), Chrome DevTools integration, and structured documentation hierarchies---all \emph{hand-crafted} to make the codebase legible to the agent. \citet{rajasekaran2026harness} report iterating through multiple rounds of evaluator prompt calibration with few-shot examples, designing four grading criteria for subjective design quality, and building a three-agent planner-generator-evaluator architecture with sprint contracts negotiated between agents. In both cases, the harness required \emph{deep domain expertise} to construct and \emph{significant iteration} to tune. \textbf{The harness improves the agent, but improving the harness still requires substantial human expertise applied to each specific task domain.} While automated prompt optimization methods such as LLM-AutoDiff \citep{yin2025llmautodiff} can tune individual components, they do not address the full harness---the tools, orchestration logic, infrastructure, and their interactions.

We propose a two-level framework that automates this improvement cycle. At the first level, the \textbf{Harness Evolution Loop} optimizes a worker agent's harness $\mathcal{H}$ for a single task through a closed-loop cycle of three agents:

\begin{enumerate}[nosep]
    \item A \textbf{Worker Agent} $W_{\mathcal{H}}$---the agent under optimization---parameterized by its harness $\mathcal{H}$, which executes a task and produces an execution trace.
    \item An \textbf{Evaluator Agent} $V$ that adversarially verifies task outcomes, diagnoses failure modes, and scores performance.
    \item An \textbf{Evolution Agent} $E$ that analyzes the full evolution history and modifies the harness---prompts, tools, orchestration logic, observations, and model configuration---to address diagnosed failure patterns.
\end{enumerate}

Starting from an initial harness $\mathcal{H}^{(0)}$---which may be a generic, untuned agent scaffold---the loop iterates for $K$ steps: at each step, the worker executes the task, the evaluator diagnoses and scores the result, and the evolution agent produces an improved harness based on the full history of prior attempts, ultimately returning the best-performing harness $\mathcal{H}^{(\text{best})}$. Together, these components form an evolution blueprint $\Lambda = (W_{\mathcal{H}}, \mathcal{H}^{(0)}, V, E)$ that fully specifies how the harness is evolved.

At the second level, a \textbf{Meta-Evolution Loop} optimizes $\Lambda$ itself across diverse tasks, learning an evolution blueprint $\Lambda^{(\text{best})}$ that enables rapid harness convergence on any new task---transforming not just harness engineering, but the design of the harness engineering process itself, into an automated optimization problem.

\begin{figure}[t]
\centering
\includegraphics[width=\textwidth]{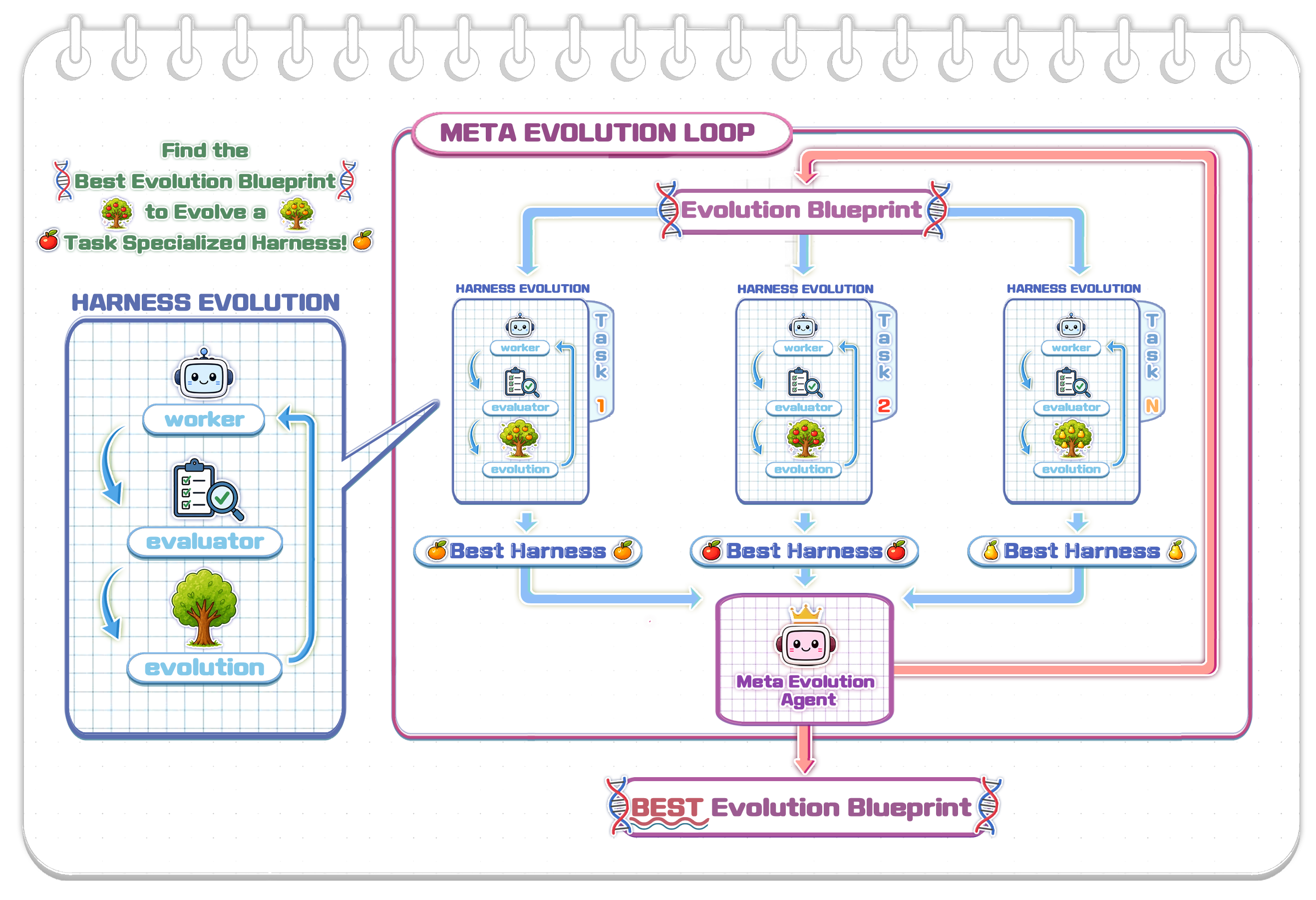}
\caption{System architecture. The \textbf{Meta-Evolution Loop} (green, outer) optimizes the evolution blueprint $\Lambda$ by running the \textbf{Harness Evolution Loop} (blue, inner) across diverse training tasks $t_1, t_2, \dots, t_n$. Each inner loop instance optimizes a worker harness $\mathcal{H}$ for a single task through iterative cycles of execution (Worker), evaluation (Evaluator), and code modification (Evolution Agent). The meta-evolution agent aggregates scores across all tasks and modifies the evolution blueprint, feeding the updated blueprint back for the next round. The output is the best-performing evolution blueprint $\Lambda^{(\text{best})}$.}
\label{fig:architecture}
\end{figure}

\section{The Harness Evolution Loop}
\label{sec:architecture}

We first formally define the notion of an agent harness, then describe the system's components---task definition, task execution, evaluation, and evolution---and how they are orchestrated in the harness evolution loop. The four components are orchestrated by a continuous automation loop, shown in Algorithm~\ref{alg:evolve-loop}.

\begin{figure}[t]
\begin{minipage}{\textwidth}
\begin{algorithm}[H]
\caption{The Harness Evolution Loop}
\label{alg:evolve-loop}
\begin{algorithmic}[1]
\REQUIRE Task $t$, Worker Agent $W_{\mathcal{H}}$, initial worker harness $\mathcal{H}^{(0)}$, Evaluator Agent $V$, Evolution Agent $E$, iterations $K$
\STATE $\mathcal{H}^{(\text{best})} \leftarrow \mathcal{H}^{(0)}$;\; $\text{best\_score} \leftarrow -\infty$
\STATE $\text{history} \leftarrow [\,]$ \COMMENT{Log of ($\mathcal{H}^{(k)}$, report, score, verdict) per iteration}
\FOR{$k = 1, 2, \dots, K$}
    \STATE \textbf{Rebuild} $W_{\mathcal{H}^{(k-1)}}$ from $\mathcal{H}^{(k-1)}$
    \STATE \textbf{Prepare} target environment; reset to clean state
    \STATE $\text{trace} \leftarrow W_{\mathcal{H}^{(k-1)}}.\text{execute}(t)$ \COMMENT{Worker runs task}
    \STATE $(\text{report}, \text{score}) \leftarrow V.\text{evaluate}(\text{trace}, t)$ \COMMENT{Evaluator diagnoses and scores}
    \IF{$\text{score} > \text{best\_score}$}
        \STATE $\text{verdict} \leftarrow \textsc{improved}$;\; $\mathcal{H}^{(\text{best})} \leftarrow \mathcal{H}^{(k-1)}$;\; $\text{best\_score} \leftarrow \text{score}$
    \ELSE
        \STATE $\text{verdict} \leftarrow \textsc{regressed}$
    \ENDIF
    \STATE $\text{history} \leftarrow \text{history} \cup \{(\mathcal{H}^{(k-1)}, \text{report}, \text{score}, \text{verdict})\}$
    \STATE $\mathcal{H}^{(k)} \leftarrow E.\text{evolve}(\text{history}, \mathcal{H}^{(\text{best})})$ \COMMENT{Evolve from best harness}
\ENDFOR
\RETURN $\mathcal{H}^{(\text{best})}, \text{best\_score}, \text{history}$
\end{algorithmic}
\end{algorithm}
\begin{algorithm}[H]
\caption{The Meta-Evolution Loop}
\label{alg:meta-evolve-loop}
\begin{algorithmic}[1]
\REQUIRE Meta-train tasks $\mathcal{T}_{\text{train}}$, Meta-Evolution Agent $E_{\text{meta}}$, initial evolution blueprint $\Lambda^{(0)}$, inner-loop budget $K$
\STATE $\Lambda^{(\text{best})} \leftarrow \Lambda^{(0)}$;\; $\text{best\_meta\_score} \leftarrow -\infty$
\STATE $\text{meta\_history} \leftarrow [\,]$
\FOR{$j = 0, 1, 2, \dots$}
    \STATE $\text{task\_results} \leftarrow [\,]$
    \FOR{each task $t_i \in \mathcal{T}_{\text{train}}$}
        \STATE $\mathcal{H}^{(\text{best})}_i, \text{best\_score}_i, \text{history}_i \leftarrow \textsc{HarnessEvolutionLoop}(t_i, \Lambda^{(j)}, K)$ \COMMENT{Alg.~\ref{alg:evolve-loop}}
        \STATE $\text{task\_results} \leftarrow \text{task\_results} \cup \{(t_i, \text{best\_score}_i, \text{history}_i)\}$
    \ENDFOR
    \STATE $\text{meta\_score} \leftarrow \text{Aggregate}(\text{task\_results})$ \COMMENT{Mean score across tasks}
    \IF{$\text{meta\_score} > \text{best\_meta\_score}$}
        \STATE $\text{verdict} \leftarrow \textsc{improved}$;\; $\Lambda^{(\text{best})} \leftarrow \Lambda^{(j)}$;\; $\text{best\_meta\_score} \leftarrow \text{meta\_score}$
    \ELSE
        \STATE $\text{verdict} \leftarrow \textsc{regressed}$
    \ENDIF
    \STATE $\text{meta\_history} \leftarrow \text{meta\_history} \cup \{(\Lambda^{(j)}, \text{task\_results}, \text{meta\_score}, \text{verdict})\}$
    \STATE $\Lambda^{(j+1)} \leftarrow E_{\text{meta}}.\text{evolve}(\text{meta\_history}, \Lambda^{(\text{best})})$ \COMMENT{Evolve from best blueprint}
\ENDFOR
\RETURN $\Lambda^{(\text{best})}, \text{best\_meta\_score}, \text{meta\_history}$
\end{algorithmic}
\end{algorithm}
\end{minipage}
\end{figure}

\subsection{Defining the Agent Harness}
\label{sec:harness-def}

A raw model is not an agent. Following \citet{trivedy2026harness}, we adopt the formulation: $\mathbf{Agent = Model + Harness}$. \textbf{A harness is every piece of code, configuration, and execution logic that is not the model itself}---it is the system that makes the model's intelligence useful. A harness can take many forms; popular categories of harness components include:
\begin{itemize}[nosep]
    \item \textbf{System prompts and task prompts}: system-level instructions that define the agent's identity and constraints, and task-level prompts that specify goals, success criteria, and in-context examples.
    \item \textbf{Tools, skills, and their descriptions}: the capabilities the agent can invoke to act on its environment (e.g., file editing, shell execution, UI interaction, web search, MCP servers).
    \item \textbf{Bundled infrastructure}: the execution environment provided to the agent (filesystem, sandboxes, browsers, observability stacks).
    \item \textbf{Orchestration logic}: the control flow that structures the agent's interaction loop (subagent spawning, handoffs, model routing, feedback loops, and continuation patterns such as the Ralph Loop).
    \item \textbf{Hooks and middleware}: deterministic execution guarantees injected around the model (compaction, continuation, lint checks, verification loops).
    \item \textbf{Model configurations}: the choice of underlying model, inference parameters (temperature, sampling strategy, token limits), and model routing rules that determine which model handles which subtask.
\end{itemize}

Harnesses appear throughout the agent ecosystem. AdaL \citep{sylphai2026adal}, Claude Code \citep{anthropic2025claudecode}, and Codex \citep{openai2025codex} are harnesses for general-purpose software engineering---they wrap LLMs with filesystem access, shell execution, web search, and multi-file editing. OpAgent \citep{guo2026opagent} is a harness for autonomous web navigation, combining a Planner, Grounder, Reflector, and Summarizer into a multi-agent pipeline that achieved state-of-the-art results on WebArena \citep{zhou2024webarena}. In every case, the harness---not the model---determines what the agent can perceive, how it acts, and how its work is orchestrated and verified.

\subsection{Task Definitions}

A task $t = (I, S)$ consists of:
\begin{itemize}[nosep]
    \item \textbf{Instructions} $I$: a concrete goal for the worker agent.
    \item \textbf{Success criteria} $S = \{s_1, s_2, \dots, s_m\}$: a checklist of verifiable conditions the evaluator uses to judge completion.
\end{itemize}

\subsection{Worker Agent}

The Worker Agent $W_{\mathcal{H}}$ is the agent under optimization---parameterized by its harness $\mathcal{H}$. It exposes a single interface $W_{\mathcal{H}}.\text{execute}(t)$: given a task $t$, the worker receives the instructions $I$, interacts with the target environment through a tool interface, and produces an execution trace $\tau$ containing environment observations, action logs, and timing information for each step.

The harness-based agents described in Section~\ref{sec:harness-def}---AdaL, Claude Code, Codex, and OpAgent---can all be set as worker agents $W_{\mathcal{H}}$, each attempting to solve a task using a specific harness configuration.

\subsection{Evaluator Agent}

The Evaluator Agent $V$ is a separate, adversarial reviewer. It exposes the interface $V.\text{evaluate}(\tau, t) \rightarrow (\text{report}, \text{score})$: given an execution trace $\tau$ and the original task $t = (I, S)$, it produces a structured diagnostic report and a numerical score. The evaluator performs four functions:

\begin{enumerate}[nosep]
    \item \textbf{State verification}: Cross-references the worker's observations in $\tau$ with ground-truth environment state to confirm the agent actually perceived what it claims, detecting hallucinated or misinterpreted states.
    \item \textbf{Criteria checking}: Evaluates the worker's final state against each success criterion $s_i \in S$, producing a pass/fail verdict per criterion.
    \item \textbf{Performance auditing}: Decomposes total execution time into \emph{LLM time} (model inference latency) versus \emph{tool time} (environment interaction latency), identifying whether bottlenecks are computational or behavioral.
    \item \textbf{Scoring}: Computes a two-tier metric---first by \emph{pass/fail} (whether the task was completed successfully), then by \emph{execution time} as a tiebreaker. This ranking determines whether a code change represents a net improvement or a regression.
\end{enumerate}

\subsection{Evolution Agent}

The Evolution Agent $E$ is the evolutionary driver of the system. It exposes the interface $E.\text{evolve}(\text{history}, \mathcal{H}^{(\text{best})}) \rightarrow \mathcal{H}'$: given the full evolution history and the best-performing harness, it produces a modified harness $\mathcal{H}'$. It operates as a senior engineer that:

\begin{enumerate}[nosep]
    \item \textbf{Aggregates diagnostics}: Reads the full evolution history---including what harness variants were tried, their evaluator reports, scores, and whether each change improved or regressed performance. This historical context prevents the evolution agent from repeating unsuccessful strategies and enables it to build on prior insights.
    \item \textbf{Identifies failure patterns}: Classifies failures into recurring categories (e.g., incorrect tool usage, reasoning loops, misinterpreted environment state, excessive latency).
    \item \textbf{Modifies the harness}: Based on the diagnosed failure patterns, the evolution agent edits the worker's harness $\mathcal{H}$---every piece of code and configuration that constitutes the agent except the model's parameters---including tool implementations, system prompts, orchestration logic, observation structure, or model configuration to address root causes.
\end{enumerate}

%
%
%

\section{Meta-Evolution: Learning to Evolve Harnesses}
\label{sec:meta-evolution}

The Harness Evolution Loop as described optimizes the worker harness $\mathcal{H}$ for a single fixed task. But the harness evolution loop itself---the evaluator prompt, the evolution agent's diagnostic strategy, the scoring function, the observation structure, and the orchestration logic---is also a harness, which we denote $\Lambda$. Formally:
\begin{equation}
    \Lambda = (W_{\mathcal{H}}, \; \mathcal{H}^{(0)}, \; V, \; E)
\end{equation}
where $W_{\mathcal{H}}$ is the worker agent, $\mathcal{H}^{(0)}$ is the initial worker harness, $V$ is the evaluator agent, and $E$ is the evolution agent. Together, these components define how the loop operates. In the current system, $\Lambda$ is designed by human engineers and remains fixed throughout the evolution process. We now describe a natural generalization: a \textbf{Meta-Evolution Agent} that optimizes $\Lambda$ itself, so that the inner harness evolution loop converges faster and more reliably to high-performing worker harnesses across diverse tasks.

\subsection{The Harness Evolution Loop as a Harness}

\textbf{Observe that $\Lambda$ has exactly the same structure as any other harness:} it consists of prompts (the evaluator and evolution agent instructions), tools (the scoring function, version control operations, code editing capabilities), observations (what telemetry and traces are surfaced from the worker, evaluator, and evolution agent), and orchestration logic (how many iterations to run, when to commit or revert, how tasks are selected and ordered). Optimizing $\Lambda$ is therefore harness optimization at a higher level of abstraction.

The components of $\Lambda$ that the Meta-Evolution Agent can modify include:
\begin{itemize}[nosep]
    \item \emph{Evaluator agent prompt}---what failure modes to look for, how to grade, what evidence to require.
    \item \emph{Evolution agent prompt}---how to diagnose failure patterns, what code changes to prioritize, how aggressively to modify the worker.
    \item \emph{Worker observation structure}---what telemetry, traces, and intermediate state to surface from the worker's execution.
    \item \emph{Evaluator and evolution agent observations}---what information flows between agents at each step.
    \item \emph{Scoring function design}---the metric structure (e.g., two-tier vs.\ multi-dimensional), thresholds, and tiebreakers.
    \item \emph{Loop hyperparameters}---number of iterations, parallelism, revert thresholds, and stopping criteria.
\end{itemize}

\subsection{A Meta-Learning Formulation}

This two-level optimization maps directly onto the meta-learning framework~\citep{thrun1998learning}. Let $\mathcal{T}_{\text{train}} = \{t_1, t_2, \dots, t_n\}$ be a set of \emph{meta-train tasks}, each representing a different agent task from potentially different domains. Let $\mathcal{T}_{\text{test}}$ be a held-out set of \emph{meta-test tasks} used to evaluate generalization.

The two loops operate as follows:
\begin{itemize}[nosep]
    \item \textbf{Inner loop} (the Harness Evolution): Given a fixed harness evolution blueprint $\Lambda$ and a single task $t_i$, run the harness evolution loop for $K$ iterations to produce an optimized worker harness $\mathcal{H}^{(K)}$. Measure the convergence trajectory: how quickly and how well the worker improves on this task.
    \item \textbf{Outer loop} (the Meta-Evolution): Across multiple tasks $t_i \in \mathcal{T}_{\text{train}}$, evaluate how effectively the current $\Lambda$ drives the inner loop. Modify $\Lambda$ to improve the \emph{speed of adaptation}---the rate at which the inner loop converges to high performance on any single task.
\end{itemize}

The objective of the outer loop is to find a harness evolution blueprint $\Lambda^{(\text{best})}$ that maximizes task performance across training tasks:
\begin{equation}
    \Lambda^{(\text{best})} = \arg\max_{\Lambda} \; \mathbb{E}_{t_i \sim \mathcal{T}_{\text{train}}} \left[ \text{best\_score}\big(\textsc{HarnessEvolutionLoop}(t_i, \Lambda, K)\big) \right]
\end{equation}
where $\textsc{HarnessEvolutionLoop}(t_i, \Lambda, K)$ runs Algorithm~\ref{alg:evolve-loop} for $K$ iterations, returning the best-performing harness $\mathcal{H}^{(\text{best})}$, its score, and the full evolution history. The evolution blueprint $\Lambda$ is judged solely by the final best score achieved on each task---not by intermediate progress.

This formulation is shown in Algorithm~\ref{alg:meta-evolve-loop}, and mirrors meta-learning, with the correspondence shown in Table~\ref{tab:meta-correspondence}.

\begin{table}[h]
\caption{Correspondence between meta-learning and meta-evolution}
\label{tab:meta-correspondence}
\begin{center}
\begin{tabular}{ll}
\multicolumn{1}{c}{\bf Meta-Learning} & \multicolumn{1}{c}{\bf Meta-Evolution} \\
\hline \\
Parameters being adapted: $\theta$ & Harness being evolved: $\mathcal{H}$ \\
Adaptation procedure ($\theta^{(0)}$, optimizer, loss) & Evolution blueprint $\Lambda = (W_{\mathcal{H}}, \mathcal{H}^{(0)}, V, E)$ \\
Inner loop: gradient updates on task $t_i$ & Inner loop: $\textsc{HarnessEvolutionLoop}(t_i, \Lambda, K)$ \\
Outer loop: meta-gradient update & Outer loop: $E_{\text{meta}}.\text{evolve}(\text{meta\_history}, \Lambda^{(\text{best})})$ \\
Meta-train tasks & Training tasks $\mathcal{T}_{\text{train}}$ \\
Meta-test tasks & Held-out tasks $\mathcal{T}_{\text{test}}$ \\
Objective: fast adaptation to new tasks & Objective: fast harness convergence on new tasks \\
\end{tabular}
\end{center}
\end{table}

\subsection{Evaluation Protocol}

Generalization is evaluated on $\mathcal{T}_{\text{test}}$: given a new, unseen task, how quickly does the inner harness evolution loop---configured with the learned $\Lambda^{(\text{best})}$---produce a high-performing worker harness? The key metrics are:
\begin{itemize}[nosep]
    \item \emph{Convergence speed}: number of inner-loop iterations to reach a target performance threshold.
    \item \emph{Final performance}: task pass rate after a fixed number of iterations.
    \item \emph{Robustness}: variance in convergence speed across different meta-test tasks.
\end{itemize}

A well-optimized $\Lambda^{(\text{best})}$ should enable the harness evolution loop to rapidly adapt to any novel task---producing effective worker harnesses with fewer iterations and less compute than a manually designed harness evolution loop.

\section{Conclusion}
\label{sec:conclusion}

We presented the \textbf{Harness Evolution Loop}, a closed-loop architecture that automatically optimizes an AI agent's harness $\mathcal{H}$---the prompts, tools, orchestration logic, and infrastructure that surround a foundation model---through repeated task execution, adversarial evaluation, and code modification. The system formalizes the agent as $W_{\mathcal{H}}$, separates evaluation ($V$) from evolution ($E$), and iteratively improves $\mathcal{H}$ while tracking the full convergence history.

We then introduced the \textbf{Meta-Evolution Loop}, which optimizes the evolution blueprint $\Lambda = (W_{\mathcal{H}}, \mathcal{H}^{(0)}, V, E)$ itself across diverse tasks. By running the inner harness evolution loop on each training task $t_i \in \mathcal{T}_{\text{train}}$ and measuring convergence, the meta-evolution agent $E_{\text{meta}}$ learns a blueprint $\Lambda^{(\text{best})}$ that enables rapid adaptation to unseen tasks. This two-level formulation mirrors meta-learning: the inner loop adapts the harness to a single task, while the outer loop optimizes the adaptation procedure for generalization.

Where harness engineering has traditionally required deep human expertise applied to each specific task domain, the Harness Evolution Loop automates this process entirely---\textbf{transforming manual harness engineering into automated harness engineering}. The Meta-Evolution Loop takes this one step further: it \textbf{automates the design of the automation itself}, learning \emph{how to evolve harnesses} rather than evolving any single harness.

We plan to follow up with empirical results on \textbf{diverse workflows that have resisted easy automation even with state-of-the-art agents and harnesses}---from complex customized customer workflows to domain-specific enterprise processes---demonstrating that the framework can crack open task categories previously considered too brittle or too specialized for autonomous agents. Ultimately, we will \textbf{release a product built on the learned evolution blueprint $\Lambda^{(\text{best})}$}: \textbf{a system where any user can point a general-purpose agent at a new task domain and have it automatically evolve into a specialized, high-performing agent}---no harness engineering expertise required.

\bibliography{iclr2026_conference}

@misc{lopopolo2026harness,
  title={Harness engineering: leveraging {Codex} in an agent-first world},
  author={Lopopolo, Ryan},
  year={2026},
  howpublished={\url{https://openai.com/index/harness-engineering/}},
  note={OpenAI Engineering Blog}
}

@misc{rajasekaran2026harness,
  title={Harness design for long-running application development},
  author={Rajasekaran, Prithvi},
  year={2026},
  howpublished={\url{https://www.anthropic.com/engineering/harness-design-long-running-apps}},
  note={Anthropic Engineering Blog}
}

@misc{trivedy2026harness,
  title={The Anatomy of an Agent Harness},
  author={Trivedy, Vivek},
  year={2026},
  howpublished={\url{https://www.langchain.com/blog/the-anatomy-of-an-agent-harness}},
  note={LangChain Blog}
}

@misc{anthropic2025claudecode,
  title={Claude Code: Best practices for agentic coding},
  author={{Anthropic}},
  year={2025},
  howpublished={\url{https://www.anthropic.com/engineering/claude-code-best-practices}},
  note={Anthropic Engineering Blog}
}

@misc{openai2025codex,
  title={Introducing {Codex}},
  author={{OpenAI}},
  year={2025},
  howpublished={\url{https://openai.com/index/introducing-codex/}},
  note={OpenAI Blog}
}

@misc{sylphai2026adal,
  title={Ada{L}: The Self-Evolving {AI} Coding Agent},
  author={{SylphAI}},
  year={2026},
  howpublished={\url{https://sylph.ai/}},
}

@inproceedings{zhou2024webarena,
  title={Web{A}rena: A Realistic Web Environment for Building Autonomous Agents},
  author={Zhou, Shuyan and Xu, Frank F and Zhu, Hao and Zhou, Xuhui and Lo, Robert and Sridhar, Abishek and Cheng, Xianyi and Ou, Tianyue and Bisk, Yonatan and Fried, Daniel and Alon, Uri and Neubig, Graham},
  booktitle={International Conference on Learning Representations (ICLR)},
  year={2024}
}

@misc{yin2025llmautodiff,
  title={{LLM-AutoDiff}: Auto-Differentiate Any {LLM} Workflow},
  author={Yin, Li and Wang, Zhangyang},
  year={2025},
  eprint={2501.16673},
  archivePrefix={arXiv},
  primaryClass={cs.CL}
}

@book{thrun1998learning,
  title={Learning to Learn},
  author={Thrun, Sebastian and Pratt, Lorien},
  year={1998},
  publisher={Springer Science \& Business Media}
}

@article{guo2026opagent,
  title={Op{A}gent: Operator Agent for Web Navigation},
  author={Guo, Yuyu and Yang, Wenjie and Yang, Siyuan and Liu, Ziyang and Chen, Cheng and Wei, Yuan and Hu, Yun and Huang, Yang and Hao, Guoliang and Yuan, Dongsheng and others},
  journal={arXiv preprint arXiv:2602.13559},
  year={2026}
}
\bibliographystyle{iclr2026_conference}

\end{document}